\documentclass[journal]{IEEEtran}

\usepackage[T1]{fontenc}
\usepackage[utf8]{inputenc}
\usepackage{amsmath,amssymb}
\usepackage{graphicx}
\usepackage[table]{xcolor}
\usepackage{booktabs}
\usepackage{multirow}
\usepackage{url}
\usepackage{cite}
\usepackage{microtype}
\usepackage{placeins}
\usepackage{tikz}
\usepackage[hidelinks]{hyperref}

\usetikzlibrary{
    arrows.meta,
    calc,
    positioning,
    decorations.pathreplacing
}

\definecolor{ArchBlue}{HTML}{EAF3FB}
\definecolor{ArchBlueLine}{HTML}{5B8DB8}
\definecolor{ArchGreen}{HTML}{EDF7EE}
\definecolor{ArchGreenLine}{HTML}{6D9B68}
\definecolor{ArchPurple}{HTML}{F2EDF8}
\definecolor{ArchPurpleLine}{HTML}{8C72AA}
\definecolor{ArchOrange}{HTML}{FFF3E8}
\definecolor{ArchOrangeLine}{HTML}{D98B3A}
\definecolor{ArchGrayLine}{HTML}{A5AAB2}
\definecolor{ArchTeal}{HTML}{2B8C91}

\newcommand{\ind}{\mathbb{I}}
\newcommand{\Fw}{F_{\beta}^{\omega}}
\newcommand{\Fmax}{F_{\beta}^{\mathrm{mx}}}
\newcommand{\MAE}{\mathcal{M}}
\newcommand{\Sm}{S_{\alpha}}
\newcommand{\Em}{E_{\phi}^{\mathrm{mn}}}
\newcommand{\LUA}{\mathcal{L}_{\mathrm{UA}}}
\newcommand{\Lrec}{\mathcal{L}_{\mathrm{rec}}}
\newcommand{\Llap}{\mathcal{L}_{\mathrm{lap}}}
\newcommand{\Lfusion}{\mathcal{L}_{\mathrm{fusion}}}

\usepackage{etoolbox}

\makeatletter

\patchcmd{\abstract}
  {\@IEEEabskeysecsize\bfseries\textit{\abstractname}}
  {\@IEEEabskeysecsize\mdseries\abstractname}
  {}{}

\patchcmd{\IEEEkeywords}
  {\@IEEEabskeysecsize\bfseries\textit{\IEEEkeywordsname}}
  {\@IEEEabskeysecsize\mdseries\IEEEkeywordsname}
  {}{}

\makeatother

\title{
MHFNet: Multi-Hypothesis Fusion Network
for Dichotomous Image Segmentation
}

\author{
Maxwell~Meyer and Jack~Spruyt\\[4pt]
\small Prama\\[2pt]
\small
\href{mailto:maxwell@backgrounderase.com}{maxwell@backgrounderase.com}
\quad
\href{mailto:jack@backgrounderase.com}{jack@backgrounderase.com}
}

\begin{document}

\maketitle

\begin{abstract}

Dichotomous image segmentation (DIS) requires accurate delineation of
foreground objects with intricate boundaries and fine structures.
Multi-hypothesis segmentation models produce alternative masks, but
standard inference selects a single candidate using an image-level
quality score, leaving the spatial information in the other candidates
unused. We introduce the Multi-Hypothesis Fusion Network (MHFNet), a
two-stage framework that refines the joint predictions of a frozen S3OD
generator. Three masks generated at $2048\times2048$ are resized to
$1024\times1024$ and concatenated with the RGB image. A hierarchical
Swin Transformer, cross-scale attention, and coarse-to-fine decoder
learn a refined mask. A mean-hypothesis confidence constraint preserves
high-confidence foreground and background assignments and restricts
corrections to intermediate-confidence pixels. On the 2,000-image
DIS-TE benchmark, MHFNet improves weighted $F$-measure from $0.8915$ to
$0.9054$ and reduces mean absolute error from $0.0277$ to $0.0255$
relative to predicted selection from the same retrained generator.
Maximum $F$-measure, structure measure, and mean enhanced-alignment
measure also improve on all four test subsets. A constraint audit
shows that $6.3\%$ of pixels are editable, while $0.84\%$ of all
pixels are incorrectly locked, exposing a limit of hard confidence
constraints. These results show consistent gains from the
complete refinement pipeline and motivate further study of
multi-hypothesis conditioning.
\end{abstract}

\begin{IEEEkeywords}
Dichotomous image segmentation,
multi-hypothesis prediction,
image segmentation,
transformer,
hypothesis fusion,
uncertainty-guided refinement,
high-resolution segmentation.
\end{IEEEkeywords}


\section{Introduction}
\label{sec:introduction}

Dichotomous image segmentation (DIS) aims to recover detailed
foreground structures from natural images, including thin components,
holes, and intricate object boundaries~\cite{Qin2022DIS}. Unlike
coarse foreground localization, DIS emphasizes pixel-level
delineation and structural completeness. Even a small segmentation
error may remove a visually important part of an object.

Recent DIS methods improve segmentation through multi-scale
representations, structural supervision, and specialized reconstruction.
InSPyReNet~\cite{Kim2022InSPyReNet}, BiRefNet~\cite{Zheng2024BiRefNet},
MVANet~\cite{Yu2024MVANet}, and PDFNet~\cite{Liu2026PDFNet} exploit
complementary feature representations to preserve detail. Generative
approaches such as GenPercept~\cite{Xu2025GenPercept} and
DiffDIS~\cite{Yu2025DiffDIS} adapt diffusion representations for
dense prediction and high-precision segmentation. These approaches
principally target the quality of the generated segmentation.

Multi-hypothesis prediction offers a complementary perspective.
Instead of generating only one interpretation, a predictor may produce
several candidate outputs~\cite{GuzmanRivera2012MCL,Rupprecht2017MHP}.
S3OD~\cite{Kupyn2026S3OD} applies this approach to foreground
segmentation through a multi-mask decoder and learned quality
estimation. At inference, the candidate with the highest predicted
quality is selected. This is a practical strategy, but a global
quality score does not describe where individual hypotheses are
correct or where they fail. Consequently, selection may discard
useful local information from the remaining masks.

We investigate whether a second-stage predictor can use the complete
hypothesis set to improve upon candidate selection. We introduce
MHFNet, which receives an RGB image and three soft masks from a frozen
S3OD generator. By processing all candidates jointly, the network can
produce an output that is not restricted to any one hypothesis.
A hierarchical transformer encoder and coarse-to-fine decoder extract
image and mask features, while a confidence constraint derived from
the mean hypothesis probability restricts learned corrections to
intermediate-confidence regions. This makes the reconstruction
problem more structured, but also prevents the network from correcting
confidently wrong assignments.

Our contributions are:
\begin{itemize}
    \item A two-stage DIS framework that replaces single-candidate
    selection with learned refinement conditioned on the RGB image
    and all three hypotheses from a frozen generator.
    \item A fusion architecture combining early image--hypothesis
    integration, hierarchical transformer encoding, cross-scale
    attention, and confidence-constrained reconstruction.
    \item A matched-generator evaluation on DIS5K, complemented by
    fixed-hypothesis and pixel-wise averaging baselines, oracle
    diagnostics, a single- versus multi-hypothesis ablation, and a
    quantitative analysis of the limits imposed by hard confidence
    locking.
\end{itemize}

On DIS-TE1--4, MHFNet consistently improves all five reported metrics
relative to the learned selector from the same retrained generator.
A single-hypothesis ablation, in which the refiner receives only the
strongest hypothesis, suggests that most of MHFNet's weighted
$F$-measure gain over that hypothesis comes from learned refinement,
while providing all three hypotheses adds a further aggregate gain
that is largest on DIS-TE4, the most complex test subset.

\section{Related Work}
\label{sec:related}

\subsection{High-Fidelity Dichotomous Segmentation}
Qin \emph{et al.}~\cite{Qin2022DIS} introduced DIS5K and IS-Net,
establishing a benchmark for detailed foreground segmentation.
InSPyReNet~\cite{Kim2022InSPyReNet} uses an inverse saliency pyramid
to combine global context with high-resolution detail.
BiRefNet~\cite{Zheng2024BiRefNet} employs bilateral reference
mechanisms for localization and reconstruction;
MVANet~\cite{Yu2024MVANet} aggregates distant and close-up views,
and PDFNet~\cite{Liu2026PDFNet} exploits pseudo-depth as a structural
prior.

Generative approaches adapt pretrained generative models to dense
prediction. GenPercept~\cite{Xu2025GenPercept} investigates the
adaptation of diffusion models to general dense perception tasks,
while DiffDIS~\cite{Yu2025DiffDIS} exploits diffusion representations
for high-precision DIS. MHFNet jointly refines multiple mask
hypotheses to improve segmentation accuracy.

\subsection{Multiple-Hypothesis Prediction}
Multiple choice learning encourages specialization across a set of
predictors by assigning training examples to suitable
hypotheses~\cite{GuzmanRivera2012MCL}. Deep multiple-hypothesis
prediction extends this principle to neural
networks~\cite{Rupprecht2017MHP}. S3OD~\cite{Kupyn2026S3OD} uses a
multi-mask decoder and a learned quality estimator for salient
object segmentation. Its learned selector returns one of the
available candidates. In contrast, MHFNet treats all candidates as
structured spatial inputs to a second-stage predictor. The
distinction is between selecting an existing mask and synthesizing
a new mask from the image and candidate set.

\subsection{Confidence-Guided Refinement}
Image matting methods commonly use a trimap to distinguish known
foreground, known background, and regions requiring estimation.
Deep Image Matting~\cite{Xu2017DeepMatting} conditions reconstruction
on the image and a trimap, while context-aware matting
\cite{Hou2019ContextMatting} exploits contextual information to
recover fine details. In DIS, BEN~\cite{Meyer2025BEN} applies
confidence-guided matting to refine uncertain regions of an initial
segmentation prediction.

MHFNet adopts a related confidence-constrained formulation for
binary foreground segmentation. Its trimap is constructed from the
mean probability of three hypotheses rather than from a single mask,
and the refinement network receives all three candidates together
with the RGB image. Thus, the focus is on hypothesis-conditioned
refinement rather than on recovering a physical alpha matte.

\section{Proposed Method}
\label{sec:method}

\subsection{Overview}

Let $\mathbf{x}$ be an RGB image and $\mathbf{y}$
its binary segmentation mask. MHFNet contains two
independently trained stages. A multi-hypothesis
generator first produces three candidate foreground
masks. A fusion network then receives the image and
all three masks to predict a refined segmentation.

\begin{figure*}[t]
\centering

\resizebox{0.99\textwidth}{!}{%
\begin{tikzpicture}[
    x=1mm,
    y=1mm,
    font=\footnotesize\sffamily,
    >=Latex,
    module/.style={
        draw,
        line width=0.65pt,
        rounded corners=1.4mm,
        align=center,
        inner xsep=2.0mm,
        inner ysep=1.4mm,
        minimum height=12mm
    },
    concat/.style={
        module,
        fill=ArchBlue,
        draw=ArchBlueLine,
        text width=18mm
    },
    stem/.style={
        module,
        fill=ArchGreen,
        draw=ArchGreenLine,
        text width=20mm
    },
    encoder/.style={
        module,
        fill=ArchPurple,
        draw=ArchPurpleLine,
        text width=19mm
    },
    attention/.style={
        module,
        fill=ArchOrange,
        draw=ArchOrangeLine,
        text width=43mm,
        minimum height=13mm
    },
    decoder/.style={
        module,
        fill=ArchGreen,
        draw=ArchGreenLine,
        text width=19mm,
        minimum height=14mm
    },
    reconstruction/.style={
        module,
        fill=ArchGreen,
        draw=ArchGreenLine,
        text width=22mm,
        minimum height=15mm
    },
    prediction/.style={
        module,
        fill=ArchBlue,
        draw=ArchBlueLine,
        text width=25mm,
        minimum height=16mm
    },
    imgbox/.style={
        draw=ArchGrayLine,
        line width=0.60pt,
        rounded corners=0.8mm,
        inner sep=0.8pt,
        fill=white
    },
    fusionimg/.style={
        imgbox,
        draw=ArchBlueLine,
        line width=0.85pt
    },
    imglabel/.style={
        draw=none,
        font=\scriptsize\sffamily,
        align=center
    },
    sectionlabel/.style={
        draw=none,
        font=\bfseries\footnotesize\sffamily,
        align=center
    },
    notelabel/.style={
        draw=none,
        font=\tiny\sffamily,
        align=center,
        text=black!65
    },
    flow/.style={
        -{Latex[length=2.0mm,width=1.25mm]},
        draw=black!78,
        line width=0.78pt
    },
    skip/.style={
        -{Latex[length=1.8mm,width=1.15mm]},
        draw=black!42,
        line width=0.62pt,
        dashed
    },
    attentionflow/.style={
        -{Latex[length=2.0mm,width=1.25mm]},
        draw=ArchOrangeLine,
        line width=0.82pt
    },
    bypass/.style={
        -{Latex[length=1.9mm,width=1.15mm]},
        draw=ArchTeal,
        line width=0.72pt,
        rounded corners=1.5mm
    }
]

\node[imgbox] (rgb) at (4,10) {
    \includegraphics[width=16.5mm,height=16.5mm,keepaspectratio]{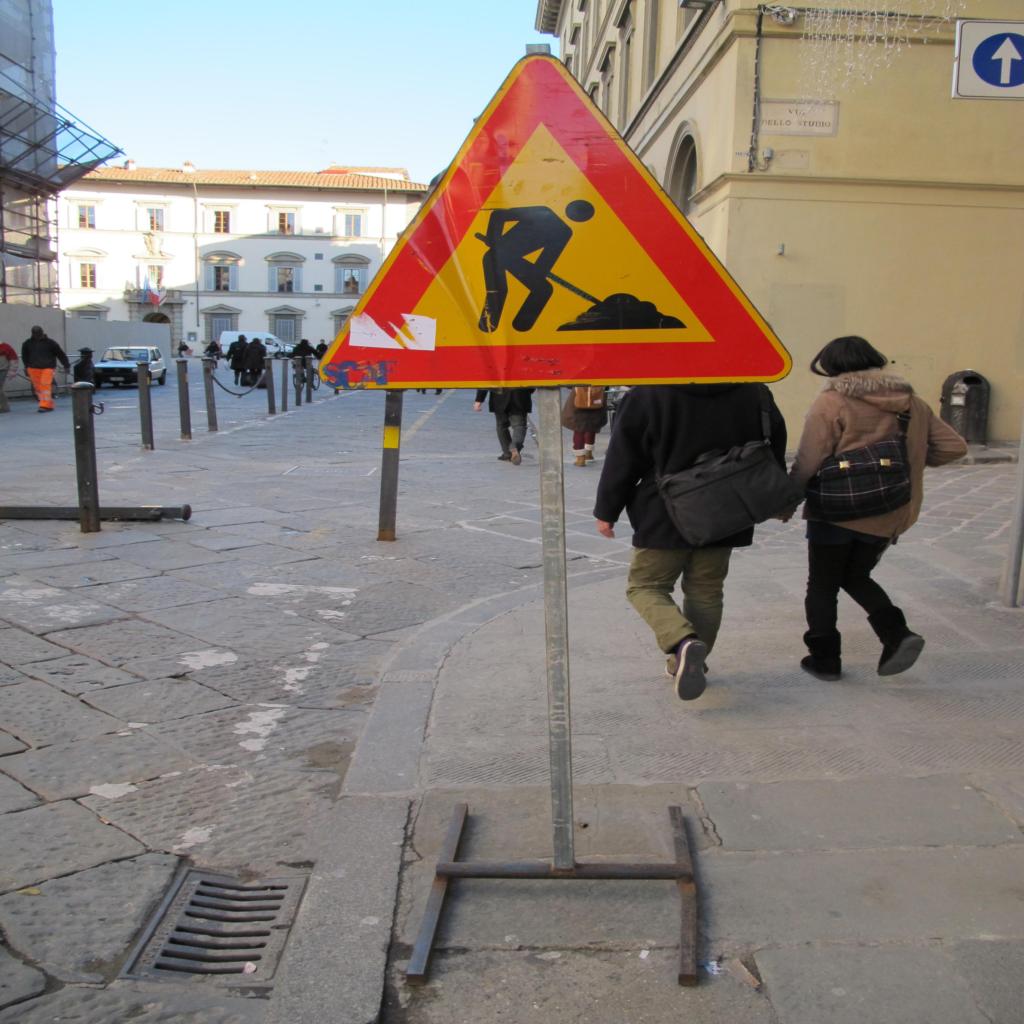}
};
\node[imglabel, above=0.8mm of rgb] {Input image \\[-0.2mm]$\mathbf{x}$};

\node[imgbox] (hyp) at (4,-14) {
    \begin{tabular}{@{}c@{\hspace{0.7mm}}c@{\hspace{0.7mm}}c@{}}
    \includegraphics[width=7.4mm,height=7.4mm,keepaspectratio]{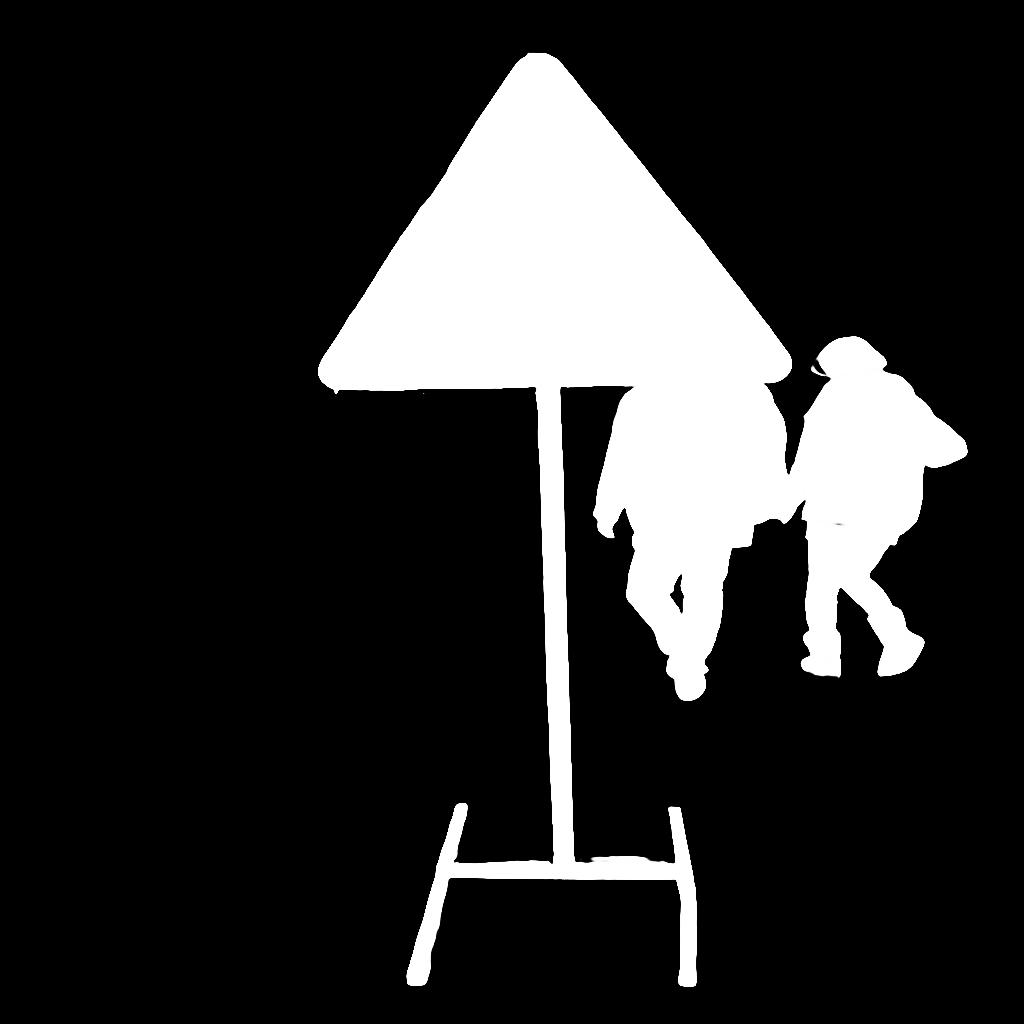}
    &
    \includegraphics[width=7.4mm,height=7.4mm,keepaspectratio]{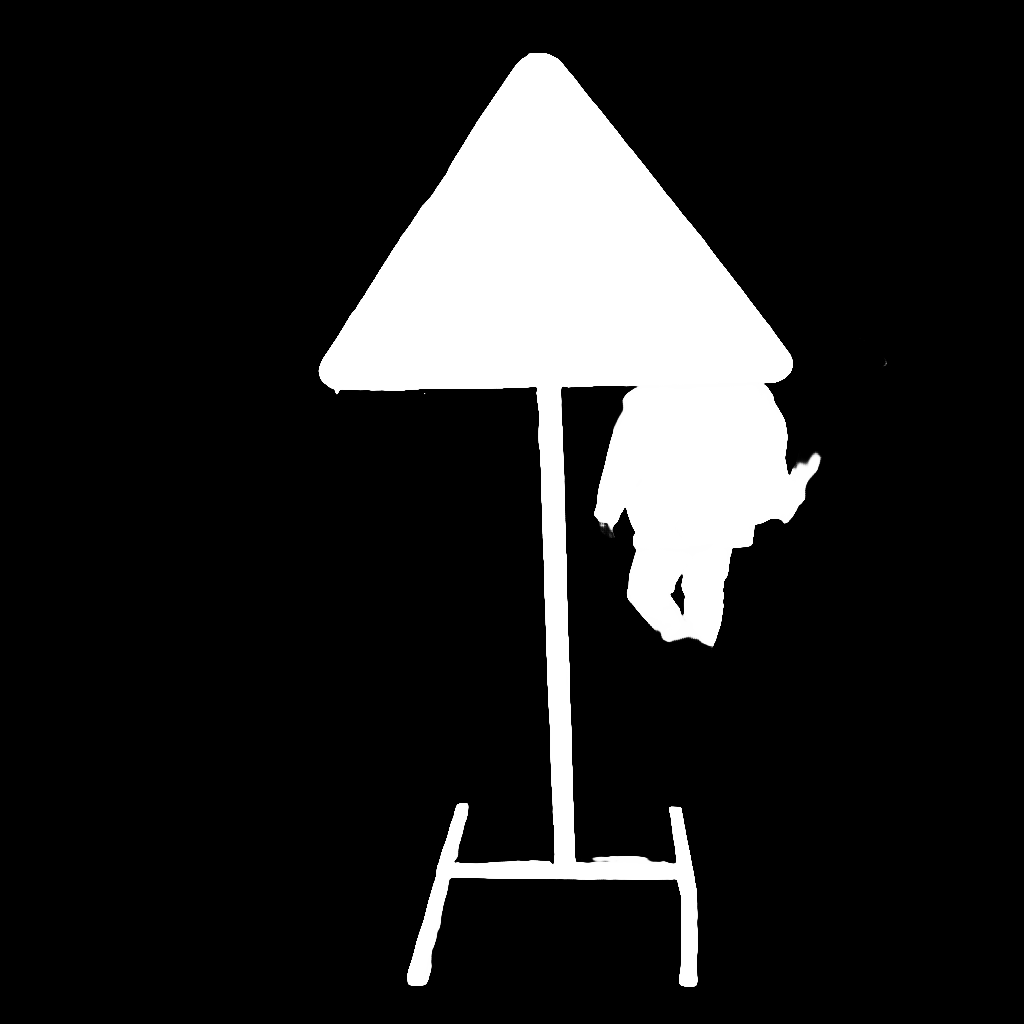}
    &
    \includegraphics[width=7.4mm,height=7.4mm,keepaspectratio]{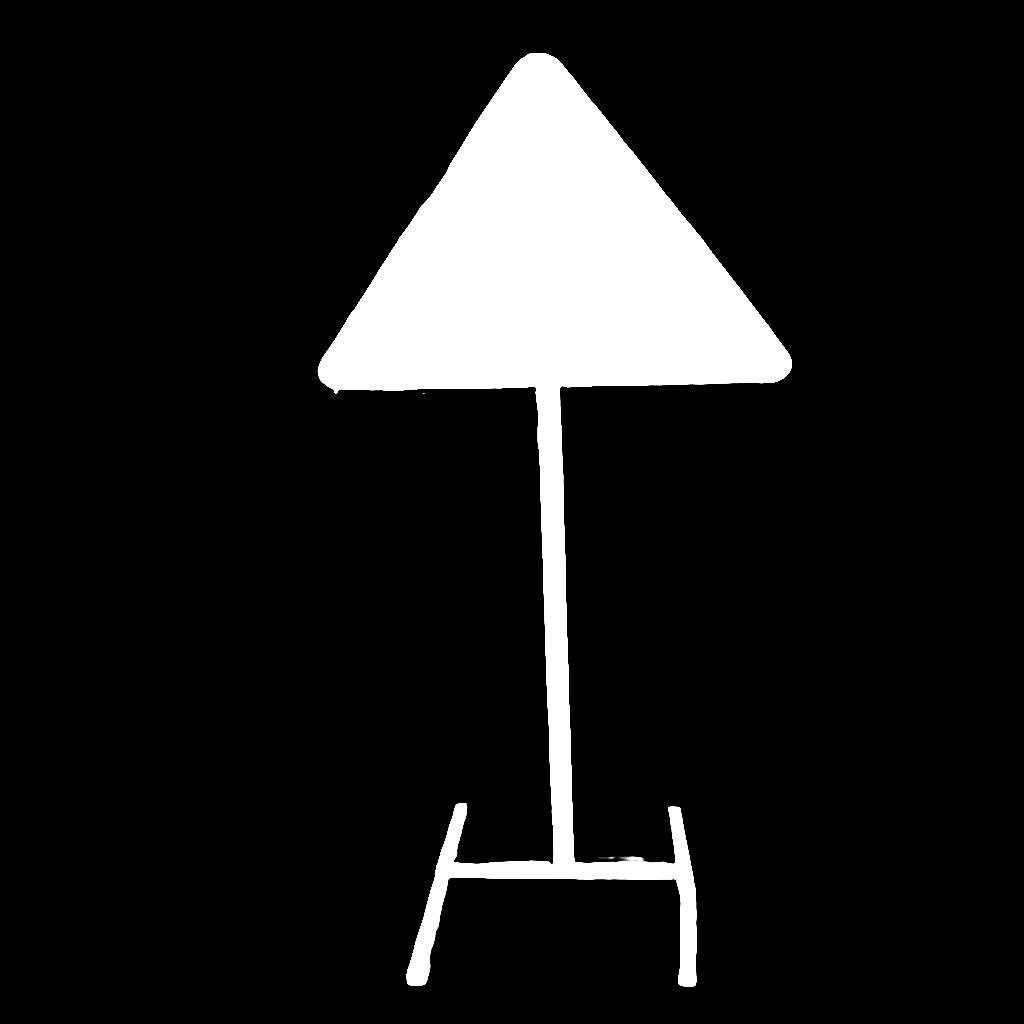}
    \\[-0.35mm]
    {\scriptsize $h_1$} & {\scriptsize $h_2$} & {\scriptsize $h_3$}
    \end{tabular}
};
\node[imglabel, above=0.8mm of hyp] {Candidate hypotheses \\[-0.2mm]$H_{\phi}(\mathbf{x})$};

\node[concat] (concat) at (29,0) {\textbf{Channel concat.}\\[0.5mm]$\mathbf{z}_0$};
\draw[flow] (rgb.east) -| ([yshift=2.6mm]concat.west);
\draw[flow] (hyp.east) -| ([yshift=-2.6mm]concat.west);

\node[stem] (shallow) at (53,0) {\textbf{Shallow-detail}\\[-0.2mm]\textbf{projection}\\[0.5mm]$\mathbf{s}_0$};
\node[stem] (embedding) at (78.5,0) {\textbf{Embedding projection}\\[-0.1mm]$+$ downsampling};
\draw[flow] (concat.east) -- (shallow.west);
\draw[flow] (shallow.east) -- (embedding.west);

\node[encoder] (x1) at (105,0) {\textbf{Swin stage 1}\\[0.6mm]$\mathbf{x}_1$};
\node[encoder] (x2) at (129,0) {\textbf{Swin stage 2}\\[0.6mm]$\mathbf{x}_2$};
\node[encoder] (x3) at (153,0) {\textbf{Swin stage 3}\\[0.6mm]$\mathbf{x}_3$};
\node[encoder] (x4) at (177,0) {\textbf{Swin stage 4}\\[0.6mm]$\mathbf{x}_4$};
\draw[flow] (embedding.east) -- (x1.west);
\draw[flow] (x1.east) -- (x2.west);
\draw[flow] (x2.east) -- (x3.west);
\draw[flow] (x3.east) -- (x4.west);


\node[attention] (attn) at (169,20) {\textbf{Cross-scale attention}\\[-0.1mm]$+$ residual fusion};
\draw[attentionflow] (x3.north) -- (x3.north |- attn.south);
\draw[attentionflow] (x4.north) -- (x4.north |- attn.south);

\node[decoder] (d4) at (177,-26) {\textbf{Decoder stage 4}\\[0.6mm]$\mathbf{d}_4$};
\node[decoder] (d3) at (149.5,-26) {\textbf{Decoder stage 3}\\[0.6mm]$\mathbf{d}_3$};
\node[decoder] (d2) at (122,-26) {\textbf{Decoder stage 2}\\[0.6mm]$\mathbf{d}_2$};
\node[decoder] (d1) at (94.5,-26) {\textbf{Decoder stage 1}\\[0.6mm]$\mathbf{d}_1$};

\draw[attentionflow]
(attn.east) -- ++(6mm,0)
|-
node[pos=0.67, right, fill=white, inner sep=0.4pt, font=\scriptsize] {$\widetilde{\mathbf{x}}_4$}
(d4.east);

\draw[flow] (d4.west) -- node[midway, above=0.8mm, fill=white, inner sep=0.45pt, font=\scriptsize] {$\mathrm{Up}_2$} (d3.east);
\draw[flow] (d3.west) -- node[midway, above=0.8mm, fill=white, inner sep=0.45pt, font=\scriptsize] {$\mathrm{Up}_2$} (d2.east);
\draw[flow] (d2.west) -- node[midway, above=0.8mm, fill=white, inner sep=0.45pt, font=\scriptsize] {$\mathrm{Up}_2$} (d1.east);

\draw[skip] (x3.south) -- (d3.north);
\draw[skip] (x2.south) -- (d2.north);
\draw[skip] (x1.south) -- (d1.north);

\draw[
    decorate,
    decoration={brace, mirror, amplitude=1.8mm},
    draw=ArchGreenLine,
    line width=0.65pt
]
([yshift=-4.0mm]d1.south west) -- ([yshift=-4.0mm]d4.south east)
node[midway, yshift=-4.8mm, sectionlabel, text=black] {Coarse-to-fine transformer decoder};

\node[reconstruction] (refine) at (67,-26) {\textbf{Full-resolution}\\[-0.2mm]\textbf{reconstruction}\\[0.5mm]$\mathbf{r}$};
\draw[flow] (d1.west) -- node[midway, above=0.8mm, fill=white, inner sep=0.45pt, font=\scriptsize] {$\mathrm{Up}_2$} (refine.east);

\node[prediction] (pred) at (38,-26) {\textbf{Prediction head}\\[0.7mm]$\mathcal{G}\!\left(\mathbf{z}_0,\mathbf{s}_0,\mathbf{r}\right)$};
\draw[flow] (refine.west) -- (pred.east);

\draw[bypass] (concat.south) -- ++(0,-10mm) -| ([xshift=-4.5mm]pred.north);
\draw[bypass] (shallow.south) -- ++(0,-7mm) -| ([xshift=4.5mm]pred.north);

\node[fusionimg] (fusion) at (18,-49) {
    \includegraphics[width=12.5mm,height=12.5mm,keepaspectratio]{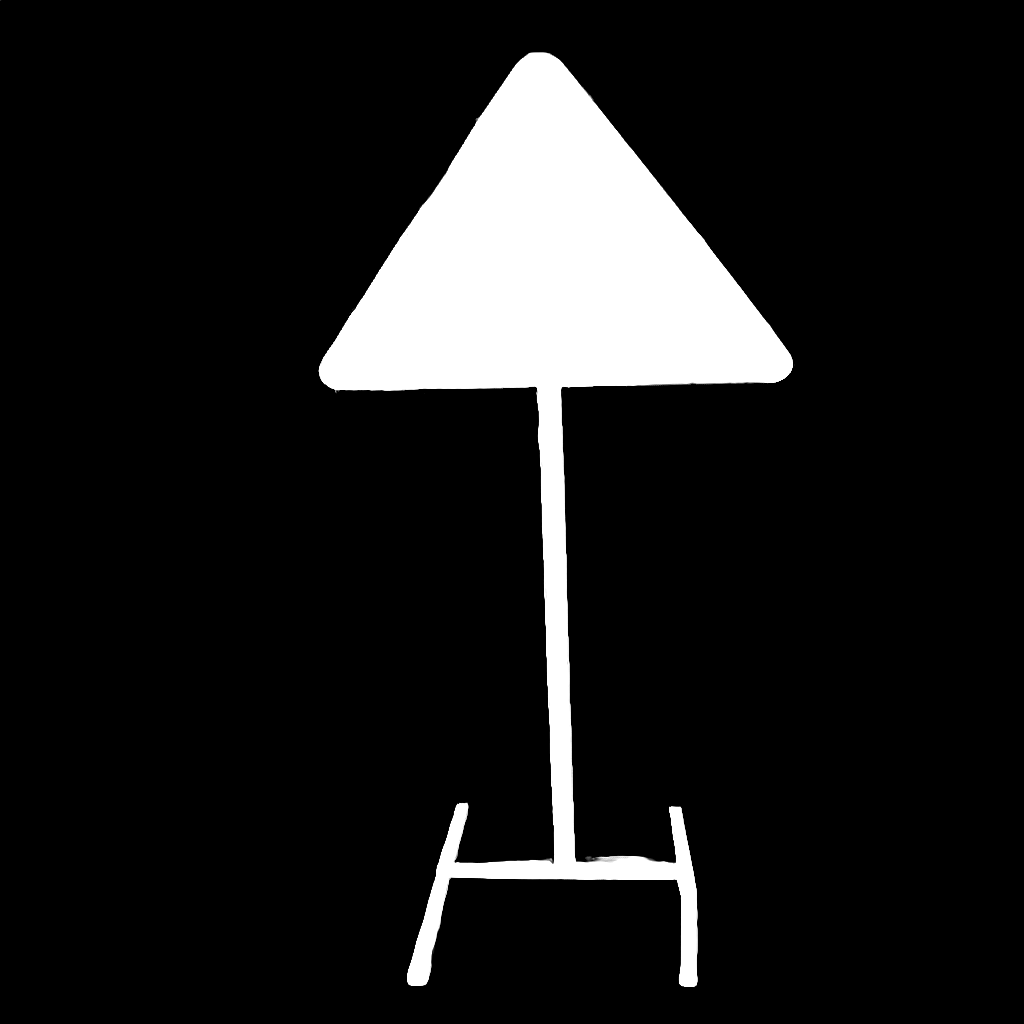}
};
\node[imgbox] (guess) at (35,-49) {
    \includegraphics[width=12.5mm,height=12.5mm,keepaspectratio]{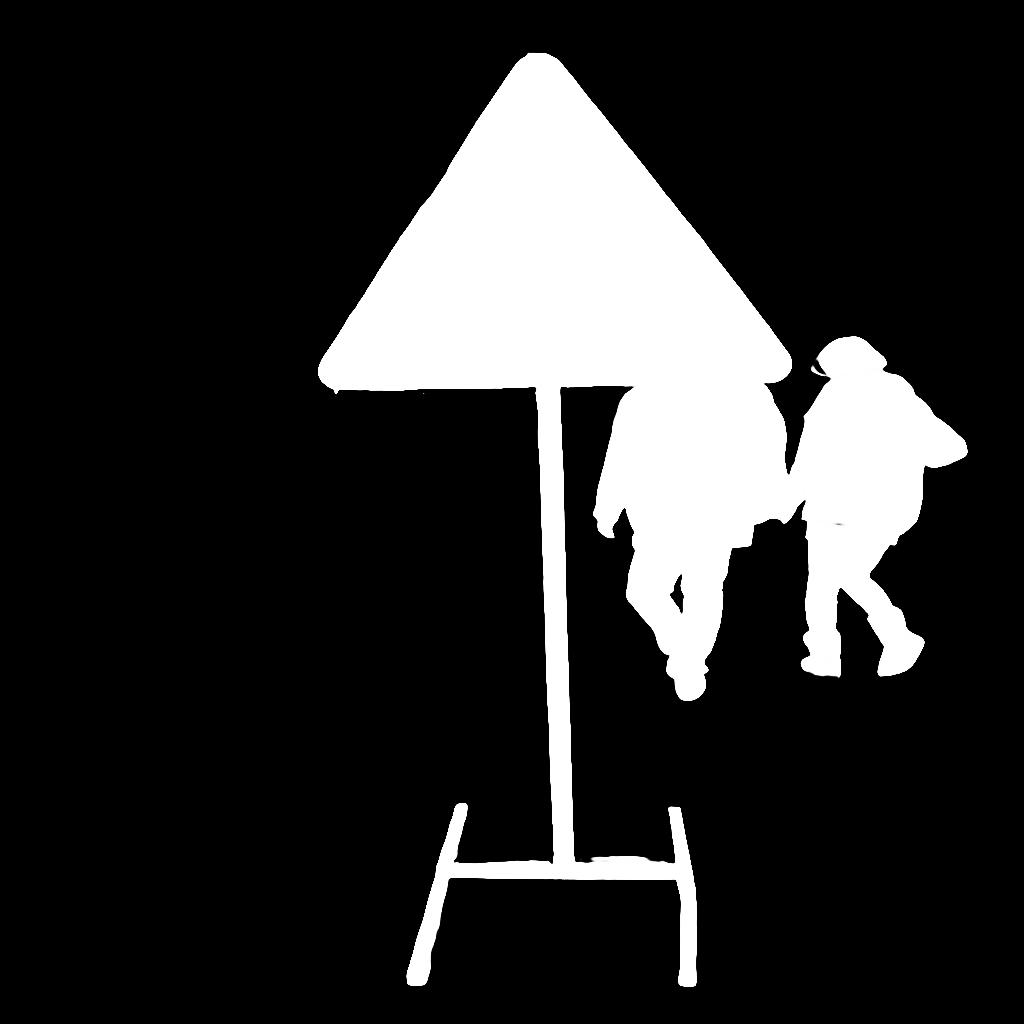}
};
\node[imgbox] (gt) at (52,-49) {
    \includegraphics[width=12.5mm,height=12.5mm,keepaspectratio]{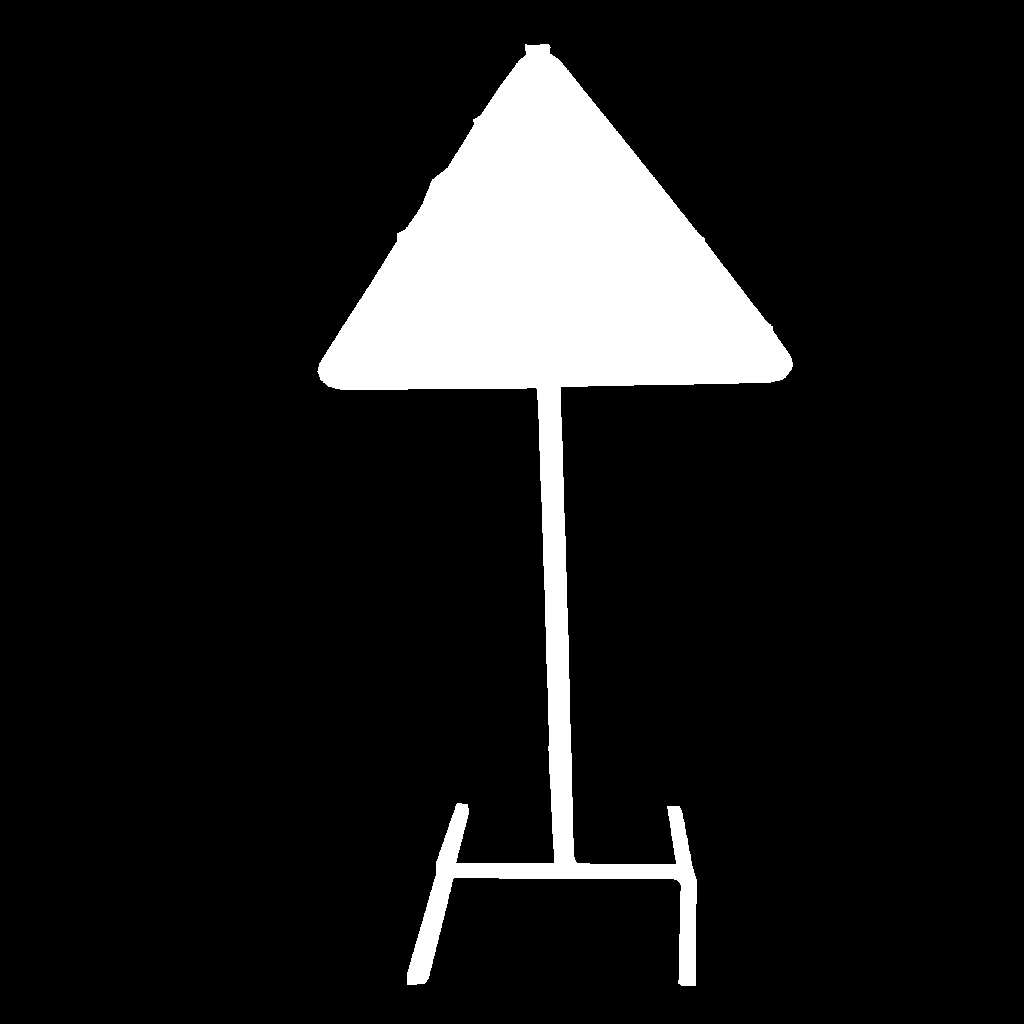}
};

\node[imglabel, below=0.7mm of fusion] {Fusion output\\[-0.2mm]$\hat{\mathbf{y}}$};
\node[imglabel, below=0.7mm of guess] {IoU-selected\\[-0.2mm]$\hat{\mathbf{y}}_{\mathrm{guess}}$};
\node[imglabel, below=0.7mm of gt] {Ground truth\\[-0.2mm]$\mathbf{y}$};
\node[notelabel, below=8.0mm of guess] {Visual comparison};

\draw[flow] (pred.south) -- ++(0,-6mm) -| (fusion.north);

\end{tikzpicture}%
}

\caption{
\textbf{Overview of the proposed hypothesis-conditioned fusion network.}
The RGB image and three candidate hypotheses are concatenated and
processed by a hierarchical Swin encoder with cross-scale attention
and a coarse-to-fine decoder. The prediction head combines
reconstructed features, shallow-detail features, and the original
input to produce the fused mask $\hat{\mathbf{y}}$.
The IoU-selected hypothesis and ground truth are shown for comparison.
}
 
\label{fig:fusion_architecture}
\end{figure*}

The generator is trained first and subsequently frozen.
Only the fusion-network parameters are optimized during
the second stage. The architecture is illustrated
in Fig.~\ref{fig:fusion_architecture}.

\subsection{Hypothesis Generation}
\label{sec:generation}

The retrained S3OD generator $H_{\phi}$ produces $K=3$ soft masks with values in $[0,1]$:

\begin{equation}
\left\{\mathbf h_k^{2K}\right\}_{k=0}^{2}
=
H_{\phi}
\left(
\mathcal R_{\mathrm{img}}^{2K}(\mathbf x)
\right).
\label{eq:generator}
\end{equation}

Here, $\mathcal R_{\mathrm{img}}^{2K}$ denotes
the generator's image preprocessing at
$2048\times2048$ resolution. The hypotheses
are resized for fusion:

\begin{equation}
\mathbf h_k
=
\mathcal R_{\mathrm{nn}}^{1K}
\left(\mathbf h_k^{2K}\right),
\qquad k=0,1,2.
\label{eq:resize}
\end{equation}

The operator $\mathcal R_{\mathrm{nn}}^{1K}$
uses nearest-neighbor interpolation to obtain
$1024\times1024$ masks. The RGB input to the
fusion network is resized separately using
bilinear interpolation.

\subsubsection{Generator Retraining}

We retain the multi-mask structure of
S3OD~\cite{Kupyn2026S3OD}, but retrain the
generator with detail-aware supervision.
The mask loss combines boundary-weighted
binary cross-entropy, weighted soft IoU, and
boundary Dice with coefficients of 4, 1, and 1.
Hypothesis assignment uses a quality target
combining soft-mask IoU (0.7) and morphological
boundary IoU (0.3).

The highest-quality hypothesis receives the
principal segmentation loss, while all branches
receive auxiliary supervision weighted by
$0.1e^{-0.2t}$ at epoch $t$. The quality head
learns these targets using MSE with weight 0.05.

Let $q_k$ denote the quality score predicted by
the generator for candidate $k$. The deployable
selection baseline is
\begin{equation}
k_{\mathrm{sel}}=\arg\max_{k\in\{0,1,2\}}q_k,
\qquad
\hat{\mathbf y}_{\mathrm{sel}}=\mathbf h_{k_{\mathrm{sel}}}.
\label{eq:predictedselection}
\end{equation}
This strategy uses no ground truth at inference.
Oracle selection instead uses ground-truth IoU
and serves only as a diagnostic.

\subsection{Hypothesis-Conditioned Fusion Architecture}
\label{sec:architecture}

The fusion network receives the
$1024\times1024$ RGB image and all three
resized hypotheses. Its input representation is

\begin{equation}
\mathbf z_0
=
\operatorname{Concat}_{c}
\left(
\mathbf x^{1K},
\mathbf h_0,
\mathbf h_1,
\mathbf h_2
\right),
\label{eq:concat}
\end{equation}

where $\mathbf x^{1K}$ is the bilinearly resized
RGB image. For three hypotheses, $\mathbf z_0$
contains six channels. This early integration
allows the network to learn joint representations
of image appearance, candidate agreement, and
hypothesis-specific errors.

\subsubsection{Convolutional Projection Stem}

A two-stage convolutional stem projects the
six-channel input into the feature space used
by the transformer backbone. The stem also
retains a shallow, high-resolution feature
representation $\mathbf s_0$ for later reconstruction.

\subsubsection{Hierarchical Encoder}

A four-stage Swin-Tiny Transformer
\cite{Liu2021Swin} extracts a hierarchy
of feature maps:

\begin{equation}
\left\{
\mathbf e_1,
\mathbf e_2,
\mathbf e_3,
\mathbf e_4
\right\}.
\label{eq:encoder}
\end{equation}

The early stages preserve more spatial detail,
while the deeper stages encode larger-context
semantic information.

\subsubsection{Cross-Scale Attention}

The two deepest encoder representations,
$\mathbf e_3$ and $\mathbf e_4$, are projected
and spatially aligned. A cross-scale attention
module uses their combined information to refine
the deepest representation into $\tilde{\mathbf e}_4$
before decoding.
This provides an interaction between the coarsest
semantic features and the preceding, more spatially
detailed feature scale.

\subsubsection{Coarse-to-Fine Decoder}

The decoder progressively upsamples the feature
representation and combines it with encoder
skip features. At decoder level $\ell$,

\begin{equation}
\begin{aligned}
\mathbf d_{\ell}
=
\mathcal B_{\ell}
\Big(
\mathcal C_{\ell}
\big(
[
\mathbf e_{\ell},
\operatorname{Up}_{2}(\mathbf d_{\ell+1})
]
\big)
\Big),
\end{aligned}
\label{eq:decoder}
\end{equation}

for $\ell\in\{3,2,1\}$. Here, $[\cdot,\cdot]$
denotes channel concatenation,
$\mathcal C_{\ell}$ is a channel projection,
and $\mathcal B_{\ell}$ is a Swin refinement
layer. The recursion starts from $\mathbf d_4$,
which decoder stage~4 computes from
$\tilde{\mathbf e}_4$.

A full-resolution reconstruction stage upsamples
$\mathbf d_1$ to produce $\mathbf r$, and the final
prediction head combines $\mathbf r$, the shallow
features $\mathbf s_0$, and the original
image--hypothesis input $\mathbf z_0$. It produces an
unconstrained probability mask
$\widetilde{\mathbf y}\in[0,1]^{1024\times1024}$.
Writing the fusion network as $G_{\theta}$, the raw prediction is
\begin{equation}
\widetilde{\mathbf y}=G_{\theta}(\mathbf z_0).
\label{eq:rawfusion}
\end{equation}

\subsection{Hypothesis-Derived Confidence Constraint}
\label{sec:constraint}

MHFNet does not directly use
$\widetilde{\mathbf y}$ as its final prediction.
Instead, the three hypotheses define a
confidence-based trimap. We first compute
their pixel-wise mean:

\begin{equation}
\overline{\mathbf h}
=
\frac{1}{3}
\sum_{k=0}^{2}
\mathbf h_k.
\label{eq:meanmask}
\end{equation}

For each pixel $p$, define

\begin{equation}
u_p
=
\ind
\left[
0.01<\overline h_p<0.99
\right]
\label{eq:unknown}
\end{equation}

and

\begin{equation}
f_p
=
\ind
\left[
\overline h_p\geq0.99
\right].
\label{eq:foreground}
\end{equation}

Pixels with $\overline h_p\leq0.01$
are assigned to confident background.
The final output is

\begin{equation}
\boxed{
\hat{\mathbf y}
=
\mathbf u
\odot
\widetilde{\mathbf y}
+
\mathbf f
}
\label{eq:constrained}
\end{equation}

where $\odot$ denotes element-wise multiplication.
Consequently,

\begin{equation}
\hat y_p
=
\begin{cases}
0,
& \overline h_p\leq0.01,\\
\widetilde y_p,
& 0.01<\overline h_p<0.99,\\
1,
& \overline h_p\geq0.99.
\end{cases}
\label{eq:piecewise}
\end{equation}
This construction restricts learned corrections
to intermediate-confidence pixels. Importantly,
the unknown region is derived from the
\emph{mean hypothesis probability}, not from
an explicit pairwise disagreement measure.
Hypotheses may agree on an intermediate
probability and still produce an unknown pixel.

The constraint is independent of ground truth
and can be applied during both training and
inference. The thresholds $0.01$ and $0.99$ are stricter than the
$0.05$/$0.95$ defaults of BEN~\cite{Meyer2025BEN}; in BEN's threshold
ablation on DIS-VD, the two settings scored equally on MAE, Dice, and
IoU. Here, the thresholds are applied to the mean of three hypotheses
rather than to a single mask. They are operational confidence rules
rather than calibrated probability guarantees and were not tuned for
MHFNet. Errors in the fixed foreground and background regions cannot
be corrected by the fusion network.

\subsection{Fusion Training Objective}
\label{sec:objective}

The fusion network is optimized using
three complementary reconstruction terms:

\begin{equation}
\boxed{
\Lfusion
=
\LUA+\Lrec+\Llap
}
\label{eq:loss}
\end{equation}

The uncertainty-aware Charbonnier term $\LUA$ uses a robust
reconstruction penalty~\cite{Charbonnier1994} to emphasize errors
inside the editable region. The global $\ell_1$ term $\Lrec$
measures full-mask reconstruction error, while the multi-scale
Laplacian-style term $\Llap$ compares high-frequency residuals at
multiple spatial resolutions, drawing on Laplacian pyramid
representations~\cite{Burt1983Laplacian}. All three coefficients
equal one in the reported implementation.

All losses are computed using the
\emph{constrained} prediction
$\hat{\mathbf y}$. Although the global
and multi-scale losses evaluate the
complete mask, their gradients with
respect to the raw prediction remain
restricted to editable pixels:

\begin{equation}
\frac{\partial\hat y_p}
{\partial\widetilde y_p}
=
u_p.
\label{eq:gradient}
\end{equation}

\section{Experimental Setup}
\label{sec:setup}

\subsection{Dataset}

We evaluate on DIS5K~\cite{Qin2022DIS},
which contains 5,470 image--mask pairs.
The dataset consists of 3,000 training
images in DIS-TR, 470 validation images
in DIS-VD, and four test subsets,
DIS-TE1 through DIS-TE4, containing
500 images each.

The test subsets span increasing levels
of object-shape and boundary complexity.
We use DIS-VD for checkpoint selection
and report results on each test subset,
their combined 2,000-image benchmark,
and DIS-VD.

\subsection{Evaluation Metrics}

Following common DIS evaluation practice,
we report weighted $F$-measure
$\Fw$~\cite{Margolin2014WeightedF},
maximum $F$-measure $\Fmax$,
mean absolute error $\MAE$,
structure measure $\Sm$~\cite{Fan2017SMeasure},
and mean enhanced-alignment measure
$\Em$~\cite{Fan2018EMeasure}.

Higher values indicate better performance
for all metrics except MAE. Following conventional saliency
evaluation~\cite{Achanta2009Saliency}, we use
$\beta^2=0.3$ for maximum $F$-measure,
with the maximum taken over the evaluated
binarization thresholds. Unless otherwise
noted, our independently evaluated masks
are scored on a common $1024\times1024$
grid. Tables display three decimal places, with ties determined at that precision. Differences stated at higher
precision are calculated from unrounded values.

\subsection{Implementation}

MHFNet adopts a two-stage architecture in which a retrained S3OD
generator produces three segmentation hypotheses at $2048\times2048$
resolution. The generator is held fixed throughout refinement training,
allowing the second stage to learn from a consistent distribution of
candidate masks. The refinement network employs a pretrained Swin-Tiny
backbone and operates at $1024\times1024$ resolution.

We train the refinement network for 100 epochs using
AdamW~\cite{Loshchilov2019AdamW} with a learning rate of $10^{-5}$.
Training is conducted on eight NVIDIA V100 GPUs with a configured
batch size of one. Model checkpoints are selected on DIS-VD, and
the four DIS-TE subsets are used to assess test performance.

\subsection{Comparison Protocol}
\label{sec:comparison}

Our experimental protocol combines literature-based benchmarking,
a matched evaluation of S3OD-based configurations, and diagnostic
analysis of the available segmentation hypotheses. This design
assesses both the competitiveness of MHFNet on DIS5K and the
improvements attributable to the proposed refinement stage.

\subsubsection{Literature Baselines}

We compare MHFNet with established DIS methods: IS-Net,
InSPyReNet, BiRefNet, GenPercept, MVANet, DiffDIS, and PDFNet-L.
Their published results are taken from Table~1 of
PDFNet~\cite{Liu2026PDFNet} and provide benchmark context for the
independently evaluated S3OD-based models.

These literature results are not independently reproduced or matched
for model capacity, pretraining, input modality, or evaluation
implementation. In particular, PDFNet notes that DiffDIS applies
pre-metric binarization, which affects metric comparability. The
primary assessment of the refinement stage therefore uses a shared
hypothesis generator and evaluation protocol.

\subsubsection{Matched Generator Evaluation}

We independently evaluate three configurations: predicted selection
from the original S3OD checkpoint, predicted selection from the
retrained 2K S3OD generator, and MHFNet operating on the hypotheses
of the retrained generator. In particular, the latter two configurations
share the same hypothesis generator and evaluation implementation,
providing a controlled assessment of replacing learned candidate
selection with the proposed refinement pipeline.

\subsubsection{Hypothesis Quality and Selection Diagnostics}

To characterize the candidate set, we score each of the three
fixed-index hypotheses independently, the pixel-wise mean of the
three soft masks without binarization, and learned candidate
selection using the predicted quality head. We also construct
best-of-three and worst-of-three oracles, which select for each image
the hypothesis with the highest or lowest IoU after binarizing the
prediction and ground truth at $0.5$. Section~\ref{sec:audit} adds a
constraint oracle that fills the editable region with ground truth,
together with an audit of the locked regions; both recompute the
trimap from the saved 8-bit hypothesis masks on the common evaluation
grid. The oracles serve exclusively as diagnostic references.
Finally, to separate the contribution of learned refinement from that
of providing all three hypotheses, we compare MHFNet with a refiner
that receives only the strongest fixed hypothesis, $h_1$
(Section~\ref{sec:hypothesis_ablation}).

\section{Results}
\label{sec:results}

\subsection{Comparison with Existing Methods}

Table~\ref{tab:overall} reports results on DIS5K. On the combined
2,000-image DIS-TE benchmark, MHFNet achieves a weighted $F$-measure
of $0.905$, maximum $F$-measure of $0.922$, MAE of $0.026$, structure
measure of $0.923$, and mean enhanced-alignment measure of $0.952$,
the best values among the methods in the table on all five metrics.
This comparison is not fully matched: MHFNet and the retrained S3OD
baseline use $2048\times2048$ generator input, whereas the published
baselines use $1024\times1024$. 

\begin{table*}[!t]
\caption{
Quantitative comparison on DIS5K across four test subsets, their
combined benchmark, and the validation set. Published baselines are
from Table~1 of PDFNet~\cite{Liu2026PDFNet}, using its PDFNet-L
variant; there, all methods were re-run with official weights at
$1024\times1024$ input. The S3OD configurations and MHFNet are
independently evaluated. S3OD (Original) uses $1024\times1024$ input,
matching the published input resolution; rows marked $^\dagger$ use
$2048\times2048$ generator input, with MHFNet refining at
$1024\times1024$. The retrained S3OD baseline uses predicted selection
from the same generator as MHFNet. DiffDIS values in the source use
pre-metric binarization and are not strictly protocol-matched. Best
and second-best distinct values are highlighted in bold and
underlined, respectively, at the displayed precision. DIS-VD is used
for checkpoint selection.
}
\label{tab:overall}
\centering
\setlength{\tabcolsep}{3.2pt}
\renewcommand{\arraystretch}{1.08}
\scriptsize
\providecommand{\tabbest}[1]{\textbf{#1}}
\providecommand{\tabsecond}[1]{\underline{#1}}
\resizebox{\textwidth}{!}{%
\begin{tabular}{@{}l|ccccc|ccccc|ccccc@{}}
\toprule
& \multicolumn{5}{c|}{\textbf{DIS-TE1 (500 images)}} & \multicolumn{5}{c|}{\textbf{DIS-TE2 (500 images)}} & \multicolumn{5}{c}{\textbf{DIS-TE3 (500 images)}} \\
\textbf{Method}
& $\Fw\!\uparrow$ & $\Fmax\!\uparrow$ & $\MAE\!\downarrow$ & $\Sm\!\uparrow$ & $\Em\!\uparrow$ & $\Fw\!\uparrow$ & $\Fmax\!\uparrow$ & $\MAE\!\downarrow$ & $\Sm\!\uparrow$ & $\Em\!\uparrow$ & $\Fw\!\uparrow$ & $\Fmax\!\uparrow$ & $\MAE\!\downarrow$ & $\Sm\!\uparrow$ & $\Em\!\uparrow$ \\
\midrule
IS-Net
 & 0.662 & 0.740 & 0.074 & 0.787 & 0.820
 & 0.728 & 0.799 & 0.070 & 0.823 & 0.858
 & 0.758 & 0.830 & 0.064 & 0.836 & 0.883 \\
InSPyReNet
 & 0.788 & 0.845 & 0.043 & 0.873 & 0.894
 & 0.846 & 0.894 & 0.036 & 0.905 & 0.916
 & 0.871 & 0.919 & 0.034 & 0.918 & 0.940 \\
BiRefNet
 & 0.819 & 0.866 & 0.037 & 0.885 & 0.911
 & 0.857 & 0.906 & 0.036 & 0.900 & 0.930
 & 0.893 & 0.920 & 0.028 & 0.919 & 0.955 \\
GenPercept
 & 0.827 & 0.850 & 0.036 & 0.878 & 0.919
 & 0.859 & 0.880 & 0.034 & 0.892 & 0.938
 & 0.879 & 0.898 & 0.032 & 0.896 & 0.954 \\
MVANet
 & 0.823 & 0.873 & 0.037 & 0.879 & 0.911
 & 0.874 & 0.916 & 0.030 & 0.915 & 0.944
 & 0.890 & 0.929 & 0.031 & 0.920 & 0.954 \\
DiffDIS
 & 0.862 & 0.883 & \tabbest{0.030} & 0.891 & \tabbest{0.933}
 & 0.895 & 0.917 & \tabsecond{0.026} & 0.913 & \tabsecond{0.951}
 & \tabsecond{0.916} & 0.934 & 0.025 & 0.919 & \tabsecond{0.964} \\
PDFNet-L
 & 0.848 & \tabsecond{0.891} & \tabbest{0.030} & 0.899 & 0.928
 & 0.886 & 0.920 & 0.027 & \tabsecond{0.924} & 0.948
 & 0.901 & 0.936 & 0.027 & 0.928 & 0.957 \\
\midrule
S3OD (Original)
 & \tabsecond{0.864} & \tabbest{0.903} & \tabsecond{0.031} & \tabsecond{0.905} & \tabsecond{0.931}
 & \tabsecond{0.900} & \tabbest{0.928} & \tabsecond{0.026} & 0.921 & \tabbest{0.952}
 & 0.907 & 0.933 & 0.026 & 0.921 & 0.959 \\
S3OD (Retrained)$^\dagger$
 & 0.856 & 0.890 & 0.032 & 0.899 & 0.927
 & 0.898 & 0.925 & 0.027 & 0.923 & 0.948
 & 0.914 & \tabsecond{0.940} & \tabsecond{0.023} & \tabsecond{0.932} & 0.961 \\
\textbf{MHFNet (Ours)}$^\dagger$
 & \tabbest{0.870} & \tabsecond{0.891} & \tabsecond{0.031} & \tabbest{0.910} & \tabbest{0.933}
 & \tabbest{0.909} & \tabsecond{0.927} & \tabbest{0.025} & \tabbest{0.926} & \tabsecond{0.951}
 & \tabbest{0.926} & \tabbest{0.942} & \tabbest{0.020} & \tabbest{0.934} & \tabbest{0.965} \\
\midrule
& \multicolumn{5}{c|}{\textbf{DIS-TE4 (500 images)}} & \multicolumn{5}{c|}{\textbf{DIS-TE1--4 (2,000 images)}} & \multicolumn{5}{c}{\textbf{DIS-VD (470 images)}} \\
\textbf{Method}
& $\Fw\!\uparrow$ & $\Fmax\!\uparrow$ & $\MAE\!\downarrow$ & $\Sm\!\uparrow$ & $\Em\!\uparrow$ & $\Fw\!\uparrow$ & $\Fmax\!\uparrow$ & $\MAE\!\downarrow$ & $\Sm\!\uparrow$ & $\Em\!\uparrow$ & $\Fw\!\uparrow$ & $\Fmax\!\uparrow$ & $\MAE\!\downarrow$ & $\Sm\!\uparrow$ & $\Em\!\uparrow$ \\
\midrule
IS-Net
 & 0.753 & 0.827 & 0.072 & 0.830 & 0.870
 & 0.725 & 0.799 & 0.070 & 0.819 & 0.858
 & 0.717 & 0.791 & 0.074 & 0.813 & 0.856 \\
InSPyReNet
 & 0.848 & 0.905 & 0.042 & 0.905 & 0.936
 & 0.838 & 0.891 & 0.039 & 0.900 & 0.922
 & 0.834 & 0.889 & 0.042 & 0.900 & 0.914 \\
BiRefNet
 & 0.864 & 0.906 & 0.039 & 0.900 & 0.939
 & 0.858 & 0.900 & 0.035 & 0.901 & 0.934
 & 0.854 & 0.897 & 0.038 & 0.898 & 0.931 \\
GenPercept
 & 0.858 & 0.874 & 0.041 & 0.874 & 0.947
 & 0.856 & 0.875 & 0.036 & 0.885 & 0.939
 & 0.859 & 0.877 & 0.035 & 0.887 & 0.941 \\
MVANet
 & 0.857 & 0.912 & 0.041 & 0.903 & 0.944
 & 0.861 & 0.908 & 0.035 & 0.904 & 0.938
 & 0.856 & 0.904 & 0.036 & 0.905 & 0.938 \\
DiffDIS
 & 0.893 & 0.909 & \tabbest{0.025} & 0.896 & \tabsecond{0.955}
 & \tabsecond{0.892} & 0.911 & \tabsecond{0.027} & 0.905 & \tabsecond{0.951}
 & 0.888 & 0.908 & 0.029 & 0.904 & 0.948 \\
PDFNet-L
 & 0.869 & 0.912 & 0.037 & 0.910 & 0.941
 & 0.876 & 0.915 & 0.030 & 0.916 & 0.944
 & 0.875 & 0.915 & 0.029 & 0.916 & 0.945 \\
\midrule
S3OD (Original)
 & 0.889 & 0.911 & 0.034 & 0.903 & 0.952
 & 0.890 & 0.919 & 0.029 & 0.912 & 0.949
 & 0.885 & 0.913 & 0.031 & 0.910 & 0.948 \\
S3OD (Retrained)$^\dagger$
 & \tabsecond{0.898} & \tabsecond{0.925} & 0.029 & \tabsecond{0.920} & \tabsecond{0.955}
 & \tabsecond{0.892} & \tabsecond{0.920} & 0.028 & \tabsecond{0.919} & 0.948
 & \tabsecond{0.894} & \tabsecond{0.922} & \tabsecond{0.027} & \tabsecond{0.923} & \tabsecond{0.953} \\
\textbf{MHFNet (Ours)}$^\dagger$
 & \tabbest{0.916} & \tabbest{0.929} & \tabsecond{0.026} & \tabbest{0.924} & \tabbest{0.960}
 & \tabbest{0.905} & \tabbest{0.922} & \tabbest{0.026} & \tabbest{0.923} & \tabbest{0.952}
 & \tabbest{0.910} & \tabbest{0.924} & \tabbest{0.024} & \tabbest{0.928} & \tabbest{0.958} \\
\bottomrule
\end{tabular}%
}
\end{table*}

Across the individual test subsets, MHFNet achieves the highest
weighted $F$-measure and structure measure on all four partitions
and leads 15 of the 20 subset--metric comparisons, counting ties at
the displayed precision. It leads all five metrics on DIS-TE3 and
four of five on DIS-TE4, where DiffDIS reports a lower MAE. On
DIS-TE1 and DIS-TE2, DiffDIS, PDFNet-L, and the original S3OD
checkpoint retain advantages on selected metrics. DIS-VD results are
shown for completeness; that split was used for checkpoint selection.

\subsection{Consistency Across DIS5K Partitions}

Against predicted selection from the same retrained generator,
MHFNet improves all five metrics on each of the four DIS-TE subsets,
on the combined test set, and on DIS-VD. On DIS-TE1--4, weighted
$F$-measure increases from $0.892$ to $0.905$ and MAE decreases from
$0.028$ to $0.026$, a relative reduction of approximately $7.9\%$
computed from unrounded scores. The largest weighted $F$-measure gain
occurs on DIS-TE4 ($0.898$ to $0.916$), and the highest absolute
score is obtained on DIS-TE3 ($0.926$). Because predicted selection
is not the strongest non-learned baseline,
Section~\ref{sec:selection} also compares MHFNet with each fixed
hypothesis and with non-learned aggregation.

\subsection{Qualitative Results}

\begin{figure*}[!t]
\centering
\includegraphics[
    width=0.88\textwidth,
    height=0.58\textheight,
    keepaspectratio
]{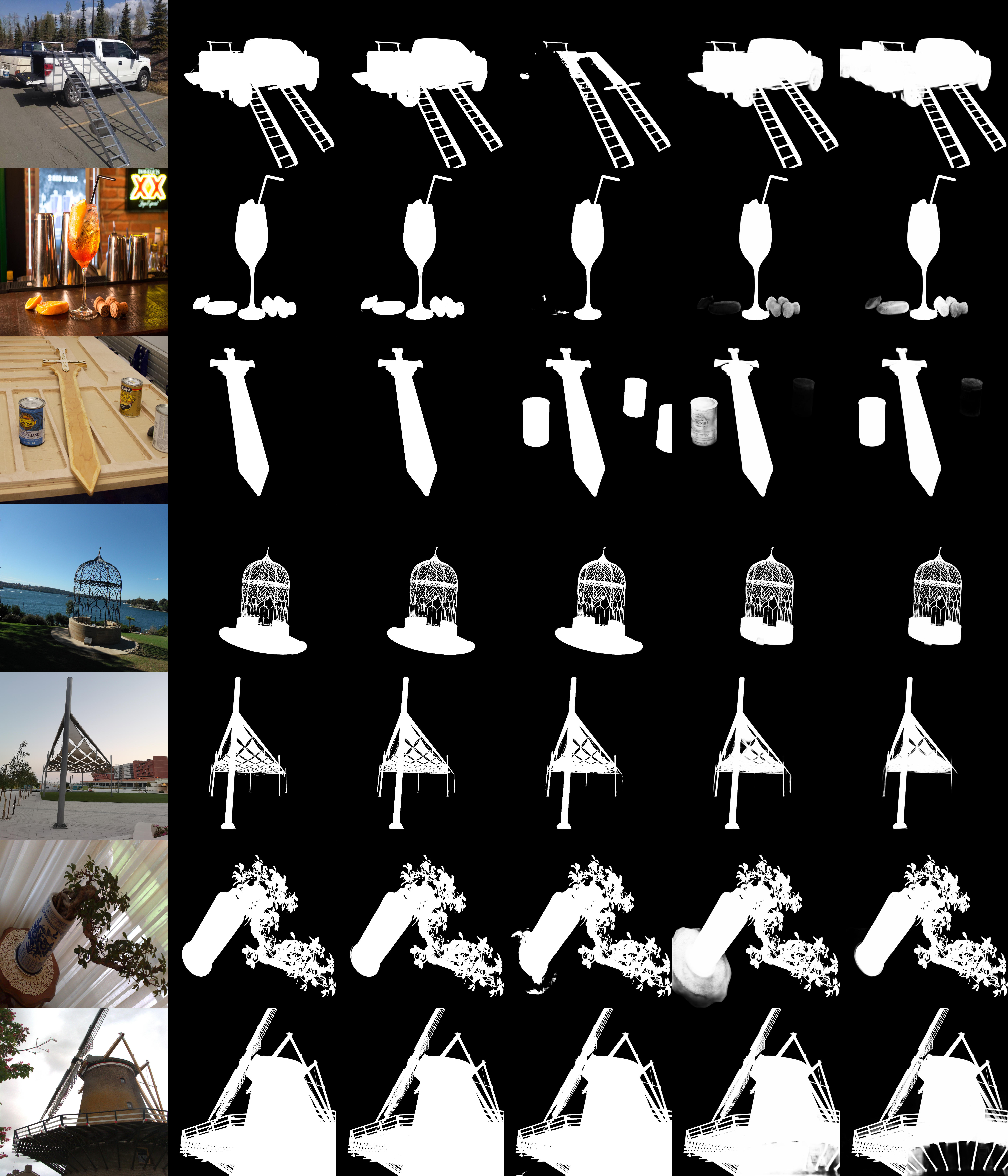}
\caption{
Qualitative comparison on DIS5K. From left to right: input image,
ground truth, MHFNet, S3OD (Original),
PDFNet, and MVANet.
}
\label{fig:qualitative}
\end{figure*}

Figure~\ref{fig:qualitative} shows representative DIS5K examples with
fine boundaries and complex foreground structures, illustrating the
spatial detail of the masks produced by MHFNet and the competing
methods.

\subsection{Hypothesis Selection Versus Fusion}
\label{sec:selection}

To assess what learned refinement adds beyond simpler uses of the
hypotheses, Table~\ref{tab:selection} compares MHFNet with
non-learned strategies built from the same generator predictions:
each fixed hypothesis, quality-head selection, and pixel-wise
averaging of the three masks. All strategies are evaluated with the
same protocol.

\begin{table}[!t]
\caption{
Non-learned strategies, MHFNet, and oracles on DIS-TE1--4 (2,000
images), using the same generator predictions and evaluation
protocol. Fixed $h_k$ returns the same hypothesis for every image.
The best- and worst-of-three oracles select candidates by
ground-truth IoU, and the constraint oracle fills editable pixels
with ground truth; oracles are not deployable. Bold marks the best
deployable result at displayed precision.
}
\label{tab:selection}
\centering
\footnotesize
\setlength{\tabcolsep}{3pt}
\renewcommand{\arraystretch}{1.10}
\resizebox{\columnwidth}{!}{%
\begin{tabular}{@{}lccccc@{}}
\toprule
\textbf{Strategy}
& $\Fw\uparrow$
& $\Fmax\uparrow$
& $\MAE\downarrow$
& $\Sm\uparrow$
& $\Em\uparrow$ \\
\midrule
Fixed $h_0$
& 0.855 & 0.884 & 0.037 & 0.900 & 0.928 \\
Fixed $h_1$
& 0.896 & 0.920 & 0.027 & \textbf{0.923} & 0.949 \\
Fixed $h_2$
& 0.893 & 0.917 & 0.029 & 0.915 & 0.946 \\
Predicted selection
& 0.892 & 0.920 & 0.028 & 0.919 & 0.948 \\
Pixel-wise mean
& 0.872 & 0.921 & 0.031 & 0.917 & 0.940 \\
\midrule
\textbf{MHFNet (Ours)}
& \textbf{0.905} & \textbf{0.922} & \textbf{0.026} & \textbf{0.923} & \textbf{0.952} \\
\midrule
Best-of-three oracle
& 0.917 & 0.936 & 0.020 & 0.936 & 0.963 \\
Worst-of-three oracle
& 0.835 & 0.869 & 0.046 & 0.884 & 0.913 \\
Constraint oracle
& 0.962 & 0.968 & 0.008 & 0.968 & 0.980 \\
\bottomrule
\end{tabular}%
}
\end{table}

The three hypotheses differ substantially in quality: $h_1$ is the
strongest on all five metrics, $h_2$ is close behind in weighted
$F$-measure ($0.893$), and $h_0$ is clearly weaker ($0.855$).
Predicted selection does not exploit this ranking. It scores below
always selecting $h_1$ on four metrics, including weighted
$F$-measure ($0.892$ versus $0.896$) and MAE ($0.0277$ versus
$0.0268$), and matches it in maximum $F$-measure. Pixel-wise
averaging is weaker still in weighted $F$-measure ($0.872$) and MAE
($0.0311$), consistent with the equal weight it gives to the weaker
$h_0$, although its maximum $F$-measure ($0.921$) exceeds that of
every fixed hypothesis.

MHFNet achieves the best value among deployable strategies on all
five metrics, tying $h_1$ in structure measure at the displayed
precision. Relative to always selecting $h_1$, it raises weighted
$F$-measure from $0.896$ to $0.905$, maximum $F$-measure from $0.920$
to $0.922$, and mean enhanced-alignment measure from $0.949$ to
$0.952$, and lowers MAE from $0.0268$ to $0.0255$. These comparisons
do not isolate individual architectural components or the effect of
the hard confidence constraint.

The ground-truth-IoU best-of-three oracle reaches weighted
$F$-measure $0.917$, indicating additional quality among the
candidate masks that neither selection nor MHFNet recovers. The
constraint oracle serves a different purpose: it replaces the
editable region with ground truth while preserving fixed
assignments, and is examined in Section~\ref{sec:audit}. Neither
oracle is deployable, and their non-IoU scores are diagnostic
measurements rather than theoretical performance bounds.

\subsection{Single-Hypothesis Versus Multi-Hypothesis Refinement}
\label{sec:hypothesis_ablation}

We next separate the contribution of the additional hypotheses by
comparing three-hypothesis MHFNet with refinement conditioned only on
the strongest fixed hypothesis, $h_1$. For the $h_1$-only refiner,
$\mathbf h_1$ fills all three hypothesis channels of $\mathbf z_0$
in~\eqref{eq:concat}, so the architecture and number of input
channels are unchanged, and the confidence constraint, computed by
the same rule, reduces to thresholding $\mathbf h_1$. The comparison
therefore measures the combined effect of the additional hypotheses
as network inputs and in the confidence constraint. Both
configurations otherwise use the same generator, training schedule,
checkpoint selection, and evaluation protocol.
Table~\ref{tab:hypothesis_ablation} reports results for the four
DIS-TE subsets, their combined benchmark, and DIS-VD; the
three-hypothesis scores are reproduced from Table~\ref{tab:overall}.

\begin{table}[!t]
\caption{
Refinement conditioned on $h_1$ alone versus all three hypotheses.
Bold marks the better value for each dataset and metric at displayed
precision; ties are bold in both rows.
}
\label{tab:hypothesis_ablation}
\centering
\footnotesize
\setlength{\tabcolsep}{2.3pt}
\renewcommand{\arraystretch}{1.08}
\resizebox{\columnwidth}{!}{%
\begin{tabular}{@{}llccccc@{}}
\toprule
\textbf{Dataset} & \textbf{Input}
& $\Fw\uparrow$
& $\Fmax\uparrow$
& $\MAE\downarrow$
& $\Sm\uparrow$
& $\Em\uparrow$ \\
\midrule
\multirow{2}{*}{DIS-TE1} & $h_1$ & 0.868 & \textbf{0.891} & \textbf{0.031} & 0.909 & 0.932 \\
 & $h_0, h_1, h_2$ & \textbf{0.870} & \textbf{0.891} & \textbf{0.031} & \textbf{0.910} & \textbf{0.933} \\
\addlinespace[2pt]
\multirow{2}{*}{DIS-TE2} & $h_1$ & \textbf{0.910} & \textbf{0.929} & \textbf{0.025} & \textbf{0.927} & \textbf{0.953} \\
 & $h_0, h_1, h_2$ & 0.909 & 0.927 & \textbf{0.025} & 0.926 & 0.951 \\
\addlinespace[2pt]
\multirow{2}{*}{DIS-TE3} & $h_1$ & 0.925 & \textbf{0.943} & 0.021 & \textbf{0.934} & 0.964 \\
 & $h_0, h_1, h_2$ & \textbf{0.926} & 0.942 & \textbf{0.020} & \textbf{0.934} & \textbf{0.965} \\
\addlinespace[2pt]
\multirow{2}{*}{DIS-TE4} & $h_1$ & 0.910 & 0.927 & 0.028 & 0.923 & 0.957 \\
 & $h_0, h_1, h_2$ & \textbf{0.916} & \textbf{0.929} & \textbf{0.026} & \textbf{0.924} & \textbf{0.960} \\
\addlinespace[2pt]
\multirow{2}{*}{DIS-TE1--4} & $h_1$ & 0.903 & \textbf{0.922} & \textbf{0.026} & \textbf{0.923} & 0.951 \\
 & $h_0, h_1, h_2$ & \textbf{0.905} & \textbf{0.922} & \textbf{0.026} & \textbf{0.923} & \textbf{0.952} \\
\addlinespace[2pt]
\multirow{2}{*}{DIS-VD} & $h_1$ & 0.908 & \textbf{0.924} & 0.025 & \textbf{0.928} & 0.957 \\
 & $h_0, h_1, h_2$ & \textbf{0.910} & \textbf{0.924} & \textbf{0.024} & \textbf{0.928} & \textbf{0.958} \\
\bottomrule
\end{tabular}%
}
\end{table}

On DIS-TE1--4, refining $h_1$ alone raises weighted $F$-measure from
$0.896$ for the unrefined hypothesis (Table~\ref{tab:selection}) to
$0.903$, and conditioning on all three hypotheses raises it further
to $0.905$. The three-hypothesis configuration also improves mean
enhanced-alignment measure from $0.951$ to $0.952$; the other three
metrics are equal at the displayed precision. The largest gains occur
on DIS-TE4, where the full configuration improves all five metrics,
and on DIS-VD it raises weighted $F$-measure from $0.908$ to $0.910$
and reduces MAE from $0.025$ to $0.024$. The benefit is not uniform:
$h_1$-only refinement performs better on four metrics on DIS-TE2 and
attains a higher maximum $F$-measure on DIS-TE3. These results
suggest that most of MHFNet's weighted $F$-measure gain over the best
fixed hypothesis comes from refinement itself, with a smaller,
subset-dependent contribution from the additional hypotheses.

\subsection{Effect of Generator Retraining}

The comparisons in Table~\ref{tab:overall} also separate the
performance associated with generator retraining from the subsequent
refinement stage. On DIS-TE1--4, the original S3OD checkpoint
achieves weighted $F$-measure $0.890$ and MAE $0.029$; the retrained
2K generator with predicted selection achieves $0.892$ and $0.028$.
Adding MHFNet to the retrained generator raises weighted $F$-measure
to $0.905$ and lowers MAE to $0.026$. Because the original and
retrained generators differ in both training objective and input
resolution, their contrast does not isolate retraining alone. The
matched retrained-generator comparison provides the more direct
evidence for the benefit of the refinement stage.

\subsection{Confidence-Constraint Analysis}
\label{sec:audit}

MHFNet uses a confidence-guided trimap to preserve high-confidence
foreground and background assignments and focus refinement on
uncertain pixels. We audit both the fraction of the image left
editable and the errors introduced by fixed assignments.

Across DIS-TE1--4, the mean-hypothesis trimap locks $93.7\%$ of
pixels and leaves $6.3\%$ editable. Of the locked pixels, $99.1\%$
agree with ground truth. The remaining incorrect assignments, about
$17.7$ million pixels or $0.84\%$ of all pixels, comprise $8.5$
million background-locked false negatives and $9.2$ million
foreground-locked false positives. These fixed errors account for
$32.3\%$ of the binary errors of the averaged prior, defined as the
mean hypothesis $\overline{\mathbf h}$ binarized at $0.5$, and cannot
be corrected without relaxing the hard constraint.

We assess the attainable correction within the editable region
using a ground-truth-filled diagnostic:
\begin{equation}
\mathbf y_{\mathrm{constraint}}^{\star}
= \mathbf u\odot\mathbf y + \mathbf f,
\label{eq:oracleconstraint}
\end{equation}
where $\mathbf u$ denotes the editable-region indicator,
$\mathbf y$ the ground truth, and $\mathbf f$ the fixed foreground
assignment. This construction preserves the locked predictions
and sets the editable pixels to their ground-truth values. For
binary ground truth, the corresponding pixel-weighted MAE floor is
\begin{equation}
\MAE_{\mathrm{floor}}
= \frac{1}{N}\sum_{p=1}^{N}(1-u_p)|f_p-y_p|,
\label{eq:maefloor}
\end{equation}
where $N$ is the number of evaluated pixels.

On DIS-TE1--4, the diagnostic increases binary IoU, averaged over
images, from $0.854$ for the averaged prior to $0.950$, and lowers
pixel-weighted binary MAE from $0.0261$ to $0.0084$, about a third of
MHFNet's MAE of $0.0255$. It also attains weighted $F$-measure
$0.962$, maximum $F$-measure $0.968$, structure measure $0.968$, and
mean enhanced-alignment measure $0.980$ (Table~\ref{tab:selection}).
The binary IoU and MAE values define the respective ceiling and floor
under the fixed assignments; the remaining metrics are empirical
diagnostic references.

\FloatBarrier

\subsection{Computational Cost}
\label{sec:computational_cost}

We evaluate computational cost on a single NVIDIA GeForce
RTX 3090 using PyTorch 2.5.1, CUDA 12.1, FP16 mixed-precision
inference, and a batch size of 1. Each configuration is
profiled over 100 measured iterations following 20 warmup
iterations. Synthetic input tensors are generated directly
on the GPU, and task-specific checkpoints are not loaded.
GPU latency is measured using synchronized CUDA events.

\begin{table}[t]
\centering
\caption{Computational cost on an NVIDIA RTX 3090 with
batch size 1 and FP16 mixed-precision inference. Latency
is the mean over 100 measured iterations; memory
denotes peak allocated GPU memory.}
\label{tab:computational_cost}
\setlength{\tabcolsep}{4pt}
\renewcommand{\arraystretch}{1.10}
\resizebox{\columnwidth}{!}{%
\begin{tabular}{lrrr}
\toprule
Configuration
& Params.
& Latency
& Memory \\
& (M)
& (ms)
& (GiB) \\
\midrule
S3OD forward
& 116.10 & 340.99 & 4.061 \\
S3OD predicted selection
& 116.10 & 343.84 & 4.109 \\
MHFNet forward
& 51.19 & 52.03 & 1.692 \\
\midrule
S3OD + MHFNet
& 167.28 & 398.10 & 4.304 \\
\bottomrule
\end{tabular}%
}
\end{table}

Table~\ref{tab:computational_cost} demonstrates that the
refinement stage introduces limited overhead relative to
high-resolution hypothesis generation. The S3OD generator
requires $340.99$ ms at $2048\times2048$ resolution, whereas
the standalone MHFNet refinement network requires only
$52.03$ ms at $1024\times1024$ resolution.

The integrated two-stage pipeline achieves a mean GPU
latency of $398.10$ ms.
Relative to the $343.84$ ms required for S3OD predicted
selection, this represents an additional latency of approximately
$54.26$ ms ($15.78\%$), with sequential throughput of
approximately $2.51$ images per second compared with $2.91$
for predicted selection.

The memory overhead is particularly modest. Despite introducing
$51.19$ million parameters and increasing the total parameter
count to $167.28$ million, the integrated pipeline requires
$4.304$ GiB of peak allocated GPU memory, compared with
$4.109$ GiB for predicted selection. This corresponds to an
increase of approximately $4.75\%$, with both networks
simultaneously resident on the GPU.

The integrated measurements encompass GPU-side normalization,
hypothesis generation, mask resizing, RGB resizing for
refinement, and application of the final confidence constraint.
Synthetic input generation, image decoding, host--device
transfers, and disk I/O are excluded.

Taken together, the accuracy and profiling results demonstrate
a favorable refinement trade-off in the evaluated GPU setting:
consistent segmentation improvements across the benchmark are
obtained with a modest increase in inference latency and
peak allocated GPU memory.

\section{Discussion}
\label{sec:discussion}

\subsection{Multi-Hypothesis Refinement Versus Selection}

The experiments support the central premise of MHFNet: the candidate
set produced by a multi-hypothesis generator contains more useful
information than any single candidate, and a learned refiner can
recover much of it without ground truth at inference. Under a shared
generator and evaluation protocol, MHFNet improves on predicted
selection on all five metrics in every DIS-TE subset and achieves the
best deployable result on all five metrics in
Table~\ref{tab:selection}, tying $h_1$ only in structure measure. It
also closes more than half of the weighted $F$-measure gap between
predicted selection and the ground-truth-IoU best-of-three oracle.

The simpler strategies help explain this gain. In this setting, the
image-level quality score is a weak basis for choosing among
candidates: predicted selection does not outperform a fixed choice of
$h_1$. Uniform averaging, in turn, gives the weaker $h_0$ the same
weight as the stronger hypotheses. MHFNet instead processes all
candidates jointly with the image, so it can draw on different
hypotheses in different regions of the same image. Most of its
weighted $F$-measure gain over the strongest fixed hypothesis comes
from learned refinement, and conditioning on all three hypotheses adds
a further aggregate gain. That gain is largest on DIS-TE4, the subset
with the most complex structures, where all five metrics improve,
suggesting that alternative hypotheses are most useful where a single
mask is most likely to miss fine detail.

\subsection{Selective Refinement Under Hard Constraints}

The mean-hypothesis constraint keeps refinement focused. It confines
learned corrections to $6.3\%$ of image pixels, while $99.1\%$ of
locked pixels already agree with ground truth, so both the
refiner's corrections and its training signal are concentrated on
uncertain regions rather than on the entire mask. The constraint
audit also makes the cost of this design explicit: incorrectly locked
pixels account for $32.3\%$ of the averaged prior's binary errors and
set an MAE floor of $0.0084$, about a third of MHFNet's MAE, because
the refiner cannot revisit them.

The constraint oracle shows that substantial headroom remains within
the current editable region. Under the same locked assignments, it
reaches weighted $F$-measure $0.962$, compared with $0.905$ for
MHFNet. Two complementary directions follow: stronger refiners can
exploit more of the editable region, and better uncertainty estimates
can reduce locked errors. Adaptive thresholds, soft constraints, and
learned confidence estimates are natural ways to balance the
preservation of reliable predictions against the recovery of missed
foreground detail.

\subsection{Scope and Future Directions}

The present study establishes the effectiveness of
hypothesis-conditioned refinement on DIS5K using a common S3OD
generator. The modular two-stage formulation offers a natural
basis for investigating transfer to alternative hypothesis
generators and segmentation datasets. Extending the evaluation
to these settings would characterize the generality of the
proposed approach beyond the current generator and benchmark.

Further investigation of confidence thresholds, individual
loss terms, architectural components, and variability across
training runs would provide additional insight into the sources
and robustness of the observed improvements. The reported
experiments use a selected checkpoint without additional
random-seed trials or confidence intervals. Literature comparisons
provide benchmark context, while matched evaluation and
profiling would be needed to establish broader accuracy and
efficiency rankings across different model families. The
integrated GPU profiling characterizes pipeline-level cost;
measurements on real image workloads, including input/output
operations and host--device transfers, would complement these
results for deployment-oriented settings.

\section{Conclusion}
\label{sec:conclusion}

We introduced MHFNet, a two-stage framework for dichotomous
image segmentation that reformulates multi-hypothesis prediction
as a confidence-guided reconstruction problem. By jointly
conditioning on the RGB image and three candidate masks from
a frozen S3OD generator, MHFNet integrates hierarchical
transformer features, cross-scale attention, and coarse-to-fine
reconstruction to produce a refined segmentation while
preserving high-confidence assignments.

Evaluation on DIS5K shows consistent
improvements over predicted hypothesis selection across all
five evaluation metrics and all four test subsets. On the
combined 2,000-image benchmark, MHFNet increases weighted
$F$-measure from $0.8915$ to $0.9054$ and reduces MAE
from $0.0277$ to $0.0255$, achieving a relative MAE reduction
of approximately $7.9\%$. It also achieves the best aggregate
values on all five metrics among the compared methods, using
$2048\times2048$ generator input rather than the
$1024\times1024$ input of the published baselines.

The confidence-constraint analysis shows that these gains are
obtained by permitting modifications within only $6.3\%$
of the output grid, while integrated profiling demonstrates
a latency increase of approximately $15.78\%$ and a peak GPU
memory increase of about $4.7\%$. Together, these findings
indicate that learned hypothesis-conditioned refinement is an
effective and computationally practical alternative to candidate
selection in the evaluated setting, and provide a foundation for
further research into selective, high-fidelity segmentation
refinement.

\bibliographystyle{IEEEtran}
\bibliography{references}

\end{document}